\documentclass[10pt,twocolumn,letterpaper]{article}

\usepackage{cvpr}
\usepackage{times}
\usepackage{epsfig}
\usepackage{graphicx}
\usepackage{amsmath}
\usepackage{amssymb}
\usepackage{hhline}
\usepackage[ruled,vlined]{algorithm2e}

\usepackage[pagebackref=true,breaklinks=true,letterpaper=true,colorlinks,bookmarks=false]{hyperref}

 \cvprfinalcopy 

\def\cvprPaperID{2593} 

\ifcvprfinal\fi
\begin{document}

\title{Minimax Defense against Gradient-based Adversarial Attacks}

\author{Blerta Lindqvist\\
Aalto University\\
{\tt\small blerta.lindqvist@aalto.fi}
\and
Rauf Izmailov\\
Perspecta Labs\\
{\tt\small rizmailov@perspectalabs.com}
}

\maketitle

\begin{abstract}
State-of-the-art adversarial attacks are aimed at neural network classifiers. By default, neural networks use gradient descent to minimize their loss function. The gradient of a classifier's loss function is used by gradient-based adversarial attacks to generate adversarially perturbed images. We pose the question whether another type of optimization could give neural network classifiers an edge. Here, we introduce a novel approach that uses minimax optimization to foil gradient-based adversarial attacks. Our minimax classifier is the discriminator of a generative adversarial network (GAN)~\cite{goodfellow2014generative} that plays a minimax game with the GAN generator. In addition, our GAN generator projects all points onto a manifold that is different from the original manifold since the original manifold might be the cause of adversarial attacks. To measure the performance of our minimax defense, we use adversarial attacks - Carlini Wagner (CW)~\cite{carlini2017towards}, DeepFool~\cite{moosavi2016deepfool}, Fast Gradient Sign Method (FGSM)~\cite{goodfellow6572explaining} - on three datasets: MNIST~\cite{lecun1998mnist}, CIFAR-10~\cite{krizhevsky2009cifar} and German Traffic Sign (TRAFFIC)~\cite{Stallkamp2012}. Against CW attacks, our minimax defense achieves 98.07\% (MNIST-default 98.93\%), 73.90\% (CIFAR-10-default 83.14\%) and 94.54\% (TRAFFIC-default 96.97\%). Against DeepFool attacks, our minimax defense achieves 98.87\% (MNIST), 76.61\% (CIFAR-10) and 94.57\% (TRAFFIC). Against FGSM attacks, we achieve 97.01\% (MNIST), 76.79\% (CIFAR-10) and 81.41\% (TRAFFIC). Our Minimax adversarial approach presents a significant shift in defense strategy for neural network classifiers.


\end{abstract}

\section{Introduction}

Machine learning classifying algorithms are susceptible to misclassification of adversarially and imperceptably perturbed inputs that are called adversarial samples. The misclassification of adversarial samples has been shown to transfer not only among diverse neural network classifiers~\cite{szegedy2013intriguing, biggio2013evasion}, but also to many other types of classifiers~\cite{szegedy2013intriguing, goodfellow6572explaining, papernot2016transferability, tramer2017space}, such as logistic regression, support vector machines, decision trees, nearest neighbors, and ensemble classifiers. One common defense against adversarial attacks on various types of classifiers is adversarial training~\cite{kurakin2016adversarial, szegedy2013intriguing, goodfellow6572explaining, tramer2017ensemble}, which augments the training data with adversarial samples. The increasing deployment of machine learning classifiers in security and safety-critical domains such as traffic signs~\cite{eykholt2018robust}, autonomous driving~\cite{amodei2016aisafety}, healthcare~\cite{faust2018deep}, and malware detection~\cite{cui2018detection} makes countering adversarial attacks important.

The field of adversarial attacks and defenses is dominated by gradient-based approaches, since gradient descent~\cite{lecun1998gradient, lecun1990handwritten} is used for optimizing neural networks. State-of-the-art adversarial attacks use the gradient of the loss function in white-box, gradient-based attacks. Such attack methods include CW~\cite{carlini2017towards}, DeepFool~\cite{moosavi2016deepfool}, FGSM~\cite{goodfellow6572explaining}, the Jacobian-based Saliency Map (JSMA)~\cite{papernot2016limitations}, the Basic Iterative Method (BIM)~\cite{kurakin2016adversarial}, ZOO~\cite{chen2017zoo}, the Projected Gradient Descent attack (PGD)~\cite{madry2017towards}. Defensive approaches against these gradient-based attacks try to mask the gradients in different ways. Defensive distillation~\cite{papernot2016distillation} does this implicitly, but not always successfully~\cite{carlini2016defensive}. Other defense methods use saturated non-linearities~\cite{nayebi2017biologically} or non-differential classifiers~\cite{lu2017safetynet} to mask gradients. However, masked and obfuscated gradients have been successfully circumvented, either by approximation of the gradient~\cite{athalye2018obfuscated}, or by using the gradients of another classifier~\cite{papernot2017practical} based on adversarial transferability~\cite{papernot2016transferability} across classifiers. The same gradient approach can also be used to circumvent defenses that use subnetworks to identify adversarial samples implicitly~\cite{meng2017magnet} or explicitly~\cite{metzen2017detecting}. Therefore, gradient-based approaches are not an effective defense for against adversarial attacks.


\begin{figure*}[t!]
\begin{center}
   \includegraphics[width=1.0\textwidth]{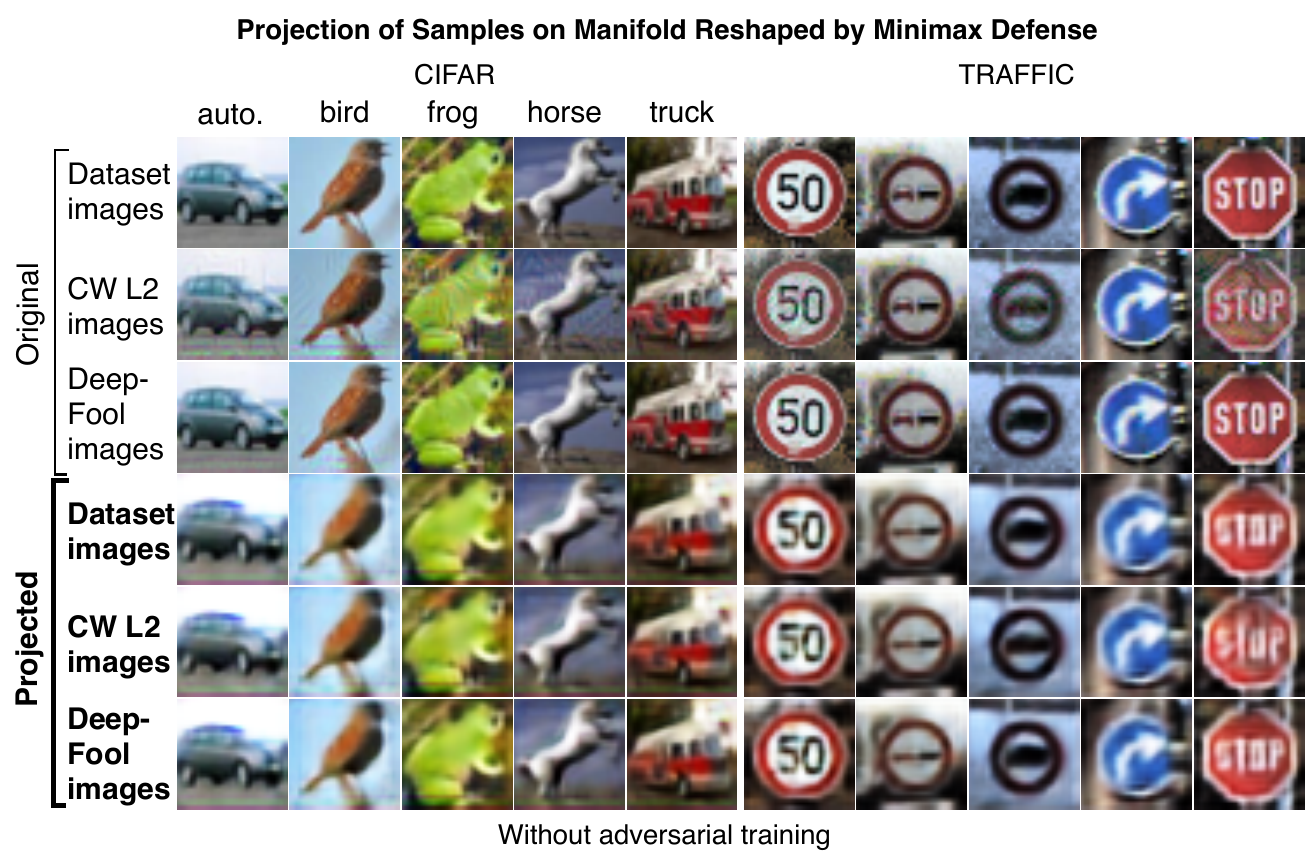}
\end{center}
   \caption{We show that original and adversarial images are projected onto a reshaped manifold, not the original one. Notice how our Minimax projections differ from the original dataset images: they are more blurry in the background and in details that do not matter for classification, for example the feathers on the bird or the window detail of the truck. Other details get highlighted. For example, the shapes of traffic signs and the signs and letters on them, the white color of the horse seems more prominent, the inside of the sign with 50 on it. For these results, our Minimax GAN has been trained only with the original training dataset, no adversarial training.}
\label{fig:fig1}
\end{figure*}

Minimax optimization by GANs has been shown~\cite{daskalakis2018limit} to reach limit points not reachable by gradient descent optimization. Whereas gradient descent~\cite{lecun1998gradient} aims to minimize a classifier's loss function, a GAN is a minimax, two-player game  between two agents, the generator and discriminator~\cite{goodfellow2014generative}. By ending up in an unreachable optimization point, a classifier that is playing a minimax game in a GAN could be able to fool gradient-based, adversarial attacks.

Many adversarial defenses aim to preserve the original manifold and probability distribution by projecting adversarial points onto the input manifold of images. In Defense-GAN~\cite{samangouei2018defense}, after random initializations in the latent space of a GAN, the closest match for the adversary image  is chosen. Both MagNet~\cite{meng2017magnet} and Ape-GAN~\cite{shen2017ape} use autoencoders to move adversarial samples towards the original manifold. PixelDefend by Song \etal~\cite{song2017pixeldefend} generates several similar images, then choses those with highest probability within a distance from an image. However, the original manifold of data points can be the cause of adversarial attacks. Low probability regions of the distribution have been attributed for the susceptibility to adversarial attacks by Szegedy \etal~\cite{szegedy2013intriguing}, Song \etal~\cite{song2017pixeldefend} and Tram{\`e}r \etal~\cite{tramer2017space}. Furthermore, adversarial attack samples have been found to be transferable to not only other deep learning classifiers of different architectures and parameters, but also to very diverse types of classifiers~\cite{papernot2016transferability}, such as logistic regression, support vector machines, decision trees, nearest neighbors, and ensemble classifiers. Such diverse classifiers have only one thing in common, the dataset, which determines the data manifold and the probability distribution. When the dataset is indeed the cause of adversarial attacks, then adhering to and maintaining the original probability distribution and manifold could be incorrect. Figure~\ref{fig:fig1} and Figure~\ref{fig:fig2} show that Minimax GAN generator does not project the images onto the original manifold.

In this paper, we present a novel defense method against state-of-the-art gradient-based adversarial attacks with two combined approaches. Based on minimax optimization in GANs, the first, novel approach counters the assumption of gradient-based attacks that neural network classifiers perform gradient descent optimization. The second approach reshapes the original manifold based on the transferability of adversarial attacks across classifiers, which points to the dataset and its manifold as the cause of adversarial attacks. The major contributions of this paper are:


\begin{itemize}
  \item A novel Minimax defense against adversarial attacks that defends from state-of-the-art gradient-based attacks.
  \item We identify current attacks as being gradient-based, on which Minimax defense is based. 
  \item Against state-of-the-art attacks, we achieve accuracy comparable to that of non-adversarial samples.
  \item To the best of our knowledge, this is the first GAN minimax approach against adversarial attacks.
\end{itemize}

\section{Related work}

\textbf{Minimax versus adversarial training.} A study on the convergence properties of gradient-based methods in minimax problems~\cite{daskalakis2018limit} has focused on their limit points, asking the question whether gradient descent methods converge to local minimax solutions. Their conclusion is that they fail to do so. This has implications for adversarial attacks and defenses, because following in this section, we will show that all current adversarial attacks are gradient-based. 

Further, we discuss adversarial attacks and defenses and provide summaries. We show that all attacks are gradient-based and that current defenses are overcome by the attacks.

\subsection{Adversarial attacks}

Here, we detail state-of-the-art adversarial attacks.






\textbf{CW} is formulated as a constrained optimization problem~\cite{carlini2017towards}:

\begin{equation} \label{eq_cw}
\begin{aligned}
& {\text{minimize}}
& &  \| x_{adv}-x_0\|_2^2 + c \cdot loss_{f}(x_{adv},l)\\
& \text{subject to}
& &  x_{adv} \in [0,1]^n,
\end{aligned}
\end{equation}
where $f$ is the classification function, $l$ is an adversary target label. With a change of variable, CW obtains an unconstrained minimization problem that allows it to do optimization through backpropagation. CW has three attacks that use the same optimization framework as in Equation \ref{eq_cw}, but are based on different norms: $L_0$, $L_2$ and $L_\infty $ attacks. CW is a gradient-based attack because it uses back-propagation on the classifier neural network.

The \textbf{DeepFool} attack~\cite{moosavi2016deepfool} looks at the distance of a point from the classifier decision boundary as the minimum amount of perturbation needed to change its classification. To estimate this distance, DeepFool approximates the classifier with a linear one, and then estimates the distance of the point from the linear boundary. After determining the minimum distance from all boundaries, DeepFool takes a step in the direction of the closest boundary. This is repeated until DeepFool finds an adversarial sample. DeepFool uses the derivative of the affine approximation of the classifier, therefore Deepfool is a gradient-based method.



The \textbf{FGSM} attack~\cite{goodfellow6572explaining} uses the gradient of a classifier's loss function with respect to the input image. FGSM perturbs each input dimension in the direction of the gradient by a magnitude of $\epsilon$. For model $\theta$, with loss $J(\theta,x,y)$, $x$ an input image and $y$ its label, adversarial images are obtained:

\begin{align}
x^{adv}=x+\epsilon sign(\nabla_x J(\theta,x,y)).
\end{align}

FGSM is a gradient-based method, because it uses the derivative. FGSM is a one-step method, its iterative version is the Basic Iterative Method (BIM)~\cite{kurakin2016adversarial}.

The \textbf{BIM} attack~\cite{kurakin2016adversarial} extends the FGSM attack~\cite{goodfellow6572explaining} by applying iteratively with a smaller step $\alpha$. After each step, pixels are clipped to keep adversarial image within $\epsilon$ neighborhood of the image. As an extension of the FGSM attack, the BIM attack is also gradient-based.


The \textbf{JSMA} attack~\cite{papernot2016limitations} is a greedy algorithm using the classifier gradient to compute a saliency map. which makes JSMA a gradient-based attack. The saliency map embodies the effect of pixels on classification. JSMA goes through the pixels in saliency-decreasing order and changes them. If misclassification is achieved the iterations are stopped.


\textbf{Universal Adversarial Perturbations (UAP)} seeks to find a universal perturbation that can cause misclassification of most points in the dataset ~\cite{moosavi2017universal}. UAP iterates over the images, calculating for each the minimal perturbation that causes that image to move to the classifier boundary. UAP aggregates all these perturbations. Several iterations over the $X$ data points are performed and the universal perturbation is centered at $0$ and its norm constrained. UAP uses the same approach as DeepFool to calculate the minimum perturbation for an image. Since DeepFool is a gradient-based attack, UAP also is gradient-based.

\textbf{Summary of adversarial attacks.} All state-of-the-art adversarial attacks are \textbf{gradient-based}. They exploit the gradients of neural network classifiers to perform optimization that creates adversarial samples. This opens up the opportunity for defending from all adversarial attacks by targeting the gradients that they relie on. CW and DeepFool are considered the strongest state-of-the-art attacks.

\subsection{Adversarial defenses}

Following are examples and details of several techniques used for defending against adversarial attacks.

\textbf{Adversarial training.} In adversarial training, the dataset is augmented with adversarial samples, often of the same kind as the attack, and the classifier is retrained~\cite{kurakin2016adversarial}. Adversarial training augments the dataset by filling out low-probability gaps with additional data points and then retraining the classifier to find a better boundary. The benefit of adversarial training is that it is easy to use and improves defense when attack is known. The drawback is that it only works against the attack that was used to generate the adversarial samples, not against others. 

\textbf{Identification or projection of adversary samples on the original manifold.} MagNet by Meng \etal~\cite{meng2017magnet} identifies adversarial samples and moves them towards the original manifold using an autoencoder or a collection of autoencoders. It contains several detector networks that detect adversarial examples from the distance between the original image and the reconstructed image. The reformer network moves adversarial samples towards the manifold. MagNet is the closest defence to our Minimax defense, but it cannot be used as a baseline for it because it has been shown~\cite{carlini2017magnet} that a small custimization of the CW attack overcomes MagNet.



Defense-GAN by Samangouei \etal~\cite{samangouei2018defense} which uses a GAN with a generator to project adversarial points onto the manifold of natural images. Given an input point that is potentially adversarial, Defense-GAN does several random initializations in the latent space and chooses from them a latent space seed that generates the closest match to the input. The closest match is considered as the projection of the original input point, though due to the randomness and depending on the number of random initializations, this projection might end up far from the real projection of the point on the manifold. Defense-GAN uses a classifier to which GAN input or GAN output or both GAN input and GAN output data points are used.

PixelDefend by Song \etal~\cite{song2017pixeldefend} proposes generative models to move adversarial images towards the distribution seen in the data. PixelDefend identifies adversarial samples with statistical methods ($p$-value) and finds more probable samples by generating similar images with an optimization that uses gradient descent, looking for highest probability images within a distance.

APE-GAN~\cite{shen2017ape} trains a pre-processing network to project normal as well as adversarial data points onto the original manifold using a GAN.

\textbf{Gradient masking and obfuscation.} Defensive Distillation (DD)~\cite{papernot2016distillation} is based on distillation - a knowledge transfer method. DD aims to provide resilience by reducing the amplitude of gradients of the loss function which are used by gradient-based attacks. DD trains a teacher network and calculates the output based on the output of the layer before the softmax layer divided by a temperature parameter T. Then, DD generates soft labels for the training dataset by running the dataset through the teacher network. The soft labels are used for training the distilled network and reduce overfitting of the original dataset.



\textbf{Summary.} Unlike adversarial training, both other defenses have been overcome by gradient-based attacks.

The approach of identification or projection of adversarial samples onto the original manifold has been successfully attacked. Identification~\cite{carlini2017adversarial} and projection on original manifold~\cite{carlini2017magnet} have been shown to be vulnerable to CW. The reason is that these defenses use neural networks to identify and to project points, and neural networks use gradient descent, which is exploited by gradient-based attacks.

Gradient masking and obfuscation has also been overcome by the CW attack~\cite{carlini2016defensive}. CW attacks the DD by changing the inputs to the final layer to avoid vanishing gradients~\cite{carlini2016defensive}. More generally, masked and \/or obfuscated gradients can be overcome due to transferability. An attack can create another classifier, train it on hard labels if available, or soft if not, and use the gradients of the new classifier for attack. Due to the transferability of adversarial attacks, adversarial attacks from the new classifier will likely transfer to the original defense.

\section{Our approach: Minimax adversarial defense}

\begin{figure*}[ht!]
\begin{center}
   \includegraphics[width=1.0\textwidth]{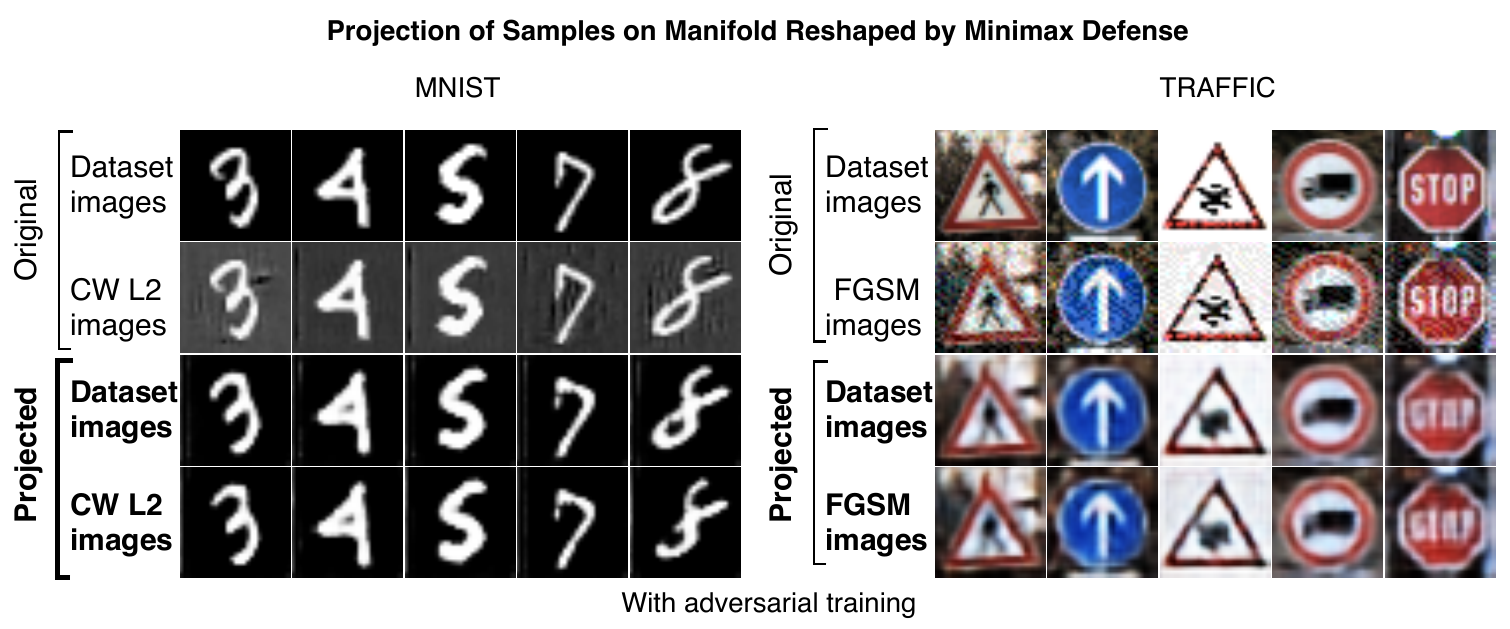}
\end{center}
   \caption{We show that MNIST and TRAFFIC images are projected on a reshaped manifold by our Minimax defense. For comparison, original images and their projections are included as well. CW and FGSM attacks are used. Notice how our Minimax projections differ from the original dataset images whether they are original images or adversary. It can be seen in the TRAFFIC images that projected images are more blurry in the background. It appears that the numbers in MNIST images and the sign details get emphasized. In the MNIST images, we can also see that the shape of the projected digits is not exactly the same as the original image. For example, digit 3 loses the little turn at the top left. Therefore, the manifold is not the same as the original one.}
\label{fig:fig2}
\end{figure*}

Our Minimax adversarial defense is a generic defense against adversarial attacks that does not identify adversarial points explicitly and does not necessitate adversarial training but can benefit from it. Figure~\ref{fig:fig2} shows image projections using the GAN generator.

Minimax defense counters state-of-the-art, gradient-based adversarial attacks by doing minimax optimization with a GAN discriminator. Our Minimax classifier is a GAN discriminator where the $Real$ and $Fake$ labels have been specialized to incorporate the classification labels. The presence of the dataset labels in the discriminator reshapes the original manifold to a different manifold. As a result, the generator projects images onto the reshaped manifold.

\textbf{What causes adversarial attacks?} It is commonly accepted that the data points of natural image datasets occupy low-dimensional manifolds in high-dimensional input space~\cite{Goodfellow-et-al-2016}. Manifolds are defined as collections of points in the input space that are connected. Many adversarial defense methods aim to preserve the original manifold and to project adversarial data points onto it. For example, Defense-GAN~\cite{samangouei2018defense} finds close matches of adversary images on the manifold and chooses the closest. MagNet~\cite{meng2017magnet} and Ape-GAN~\cite{shen2017ape} move points onto the original manifold with autoencoder and GAN autoencoder generator respectively. PixelDefend by Song \etal~\cite{song2017pixeldefend} chooses a substitute for an adversarial image by generating several similar images, then choosing those with highest probability within a distance from the adversary image. We argue that the adversarial transferability across very different classifiers~\cite{papernot2016transferability} means that the dataset and its manifold shape are the cause, specifically low probability regions of the distribution~\cite{szegedy2013intriguing,song2017pixeldefend,tramer2017space}. Furthermore, adversarial attack samples have been found to be transferable to not only other deep learning classifiers of different architectures and parameters, but also to very diverse types of classifiers~\cite{papernot2016transferability}, such as logistic regression, support vector machines, decision trees, nearest neighbors, and ensemble classifiers. Such diverse classifiers have only one thing in common, the dataset, which determines the data manifold and the probability distribution. Therefore, to counter adversarial attacks, the manifold needs to be reshaped and not preserved.

\begin{figure}[t!]
\begin{center}
   \includegraphics[width=0.4\textwidth]{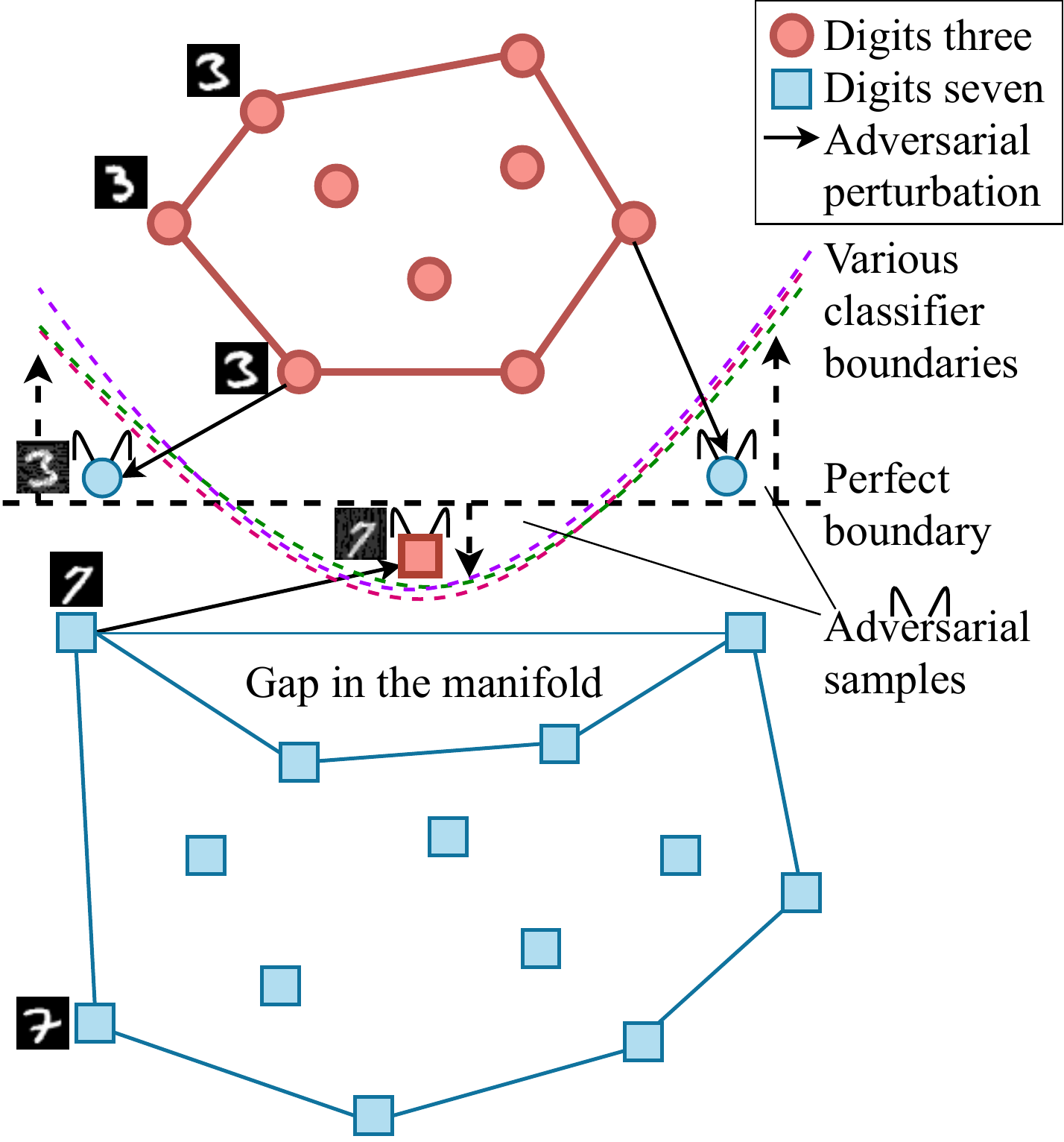}
\end{center}
   \caption{We illustrate our understanding of how manifold gaps cause adversarial attacks for a simple case of classification of digits $3$ and $7$. There is a small gap in the lower manifold that lacks data points from the square class. Classifiers aim to maintain same distance from points of different classes. This causes them to pass roughly in the same region of space. Due to the manifold gap, all classifiers shift towards the gap to maintain same distance from points of different classes. As a result, in the vicinity of the gap, all classifiers are surrounded by space the points of which would belong to one of the classes. They all misclassify the data points shown as adversarial samples, which are between the classifiers and the perfect boundary.}
\label{fig:adv_pockets}
\end{figure}

Figure~\ref{fig:adv_pockets} depicts our view that low probability regions cause adversarial attacks~\cite{szegedy2013intriguing,song2017pixeldefend,tramer2017space}. The boundaries of different types of classifiers overlap each-other, as supported by adversarial transferability. As all classifiers aim to keep same distances from different dataset classes, gaps in the manifold cause the boundaries to be shifted. Shifted boundaries create areas in which images will misclassify due to the shift.

\textbf{GAN choice} To reshape manifolds, autoencoders are frequently used, either on their own~\cite{meng2017magnet} as in MagNet~\cite{meng2017magnet}, or as part of a GAN with an autoencoder generator~\cite{zhu2016generative}. We choose to use a GAN in our approach because GANs have been shown to be very good at reconstructing manifolds of natural image datasets in original high-dimensional input spaces~\cite{goodfellow2014generative}. In distinction to other methods that also use GANs against adversarial attacks, our Minimax generator is an autoencoder because it enables direct projection of images onto the reshaped manifold. The inclusion of class labels into $Real$ and $Fake$ labels of our Minimax discriminator allows our defense to reshape the manifold even in the absence of adversarial samples in training.

\textbf{Multi-label GAN discriminator in Minimax adversarial defense.} In Minimax defense, we combine and extend these GAN variations~\cite{odena2016conditional, salimans2016improved}. We combine them by having the GAN discriminator act as a classifier and extend them by having $K$ labels for each $Real$ and $Fake$ category of labels. As a result, a Minimax discriminator has twice the number of labels in the dataset, in all $2 \times K$ labels.

\textbf{How our approach differs from similar defenses.} The defenses that are most similar to Minimax defense, MagNet~\cite{meng2017magnet} and Defense-GAN~\cite{samangouei2018defense}, are essentially different from Minimax in that they do not deply the minimax defense. MagNet~\cite{meng2017magnet} and Defense-GAN~\cite{samangouei2018defense} reposition their sample points with autoencoders/GANs and then classify the repositioned points on a separate classifier. Our Minimax defense uses the GAN discriminator as classifier, does not perform gradient descent. Instead, the descriminator/classifier is involved in a minimax game with the GAN generator, which leads it to different optimization solutions than gradient descent. This is the crucial difference from MagNet~\cite{meng2017magnet} and Defense-GAN~\cite{samangouei2018defense}.

\textbf{Labels.} The original GAN definition~\cite{goodfellow2014generative} specified two discriminator labels: $Real$, $Fake$. Our Minimax defense further breaks down these labels into the classes of the classification problem. If the original classification problem has $K$ classes, our Minimax defense discriminator has $2 \times K$ classes. For example, for MNIST, the discriminator labels are: $Real-0$, $Fake-0$, $Real-1$, $FAKE-1$, \ldots\ , $REAL-9$ and $FAKE-9$; 20 labels in total.

\textbf{Loss function.} The Minimax GAN loss function is:
\begin{multline}
L = \sum_{i=0}^{9} E_{x \sim\ p_{data}(x) | y_{i}}[log D(x)] + \\ + \sum_{i=10}^{19} E_{x \sim\ p_{data}(x) | y_{i}}[1-log D(G(x))].
\end{multline}

The term $log D(G(x))$ reflects the usage of an autoencoder as the generator, where the input is dataset images.

\begin{figure*}[t]
\begin{center}
  \includegraphics[width=0.65\linewidth]{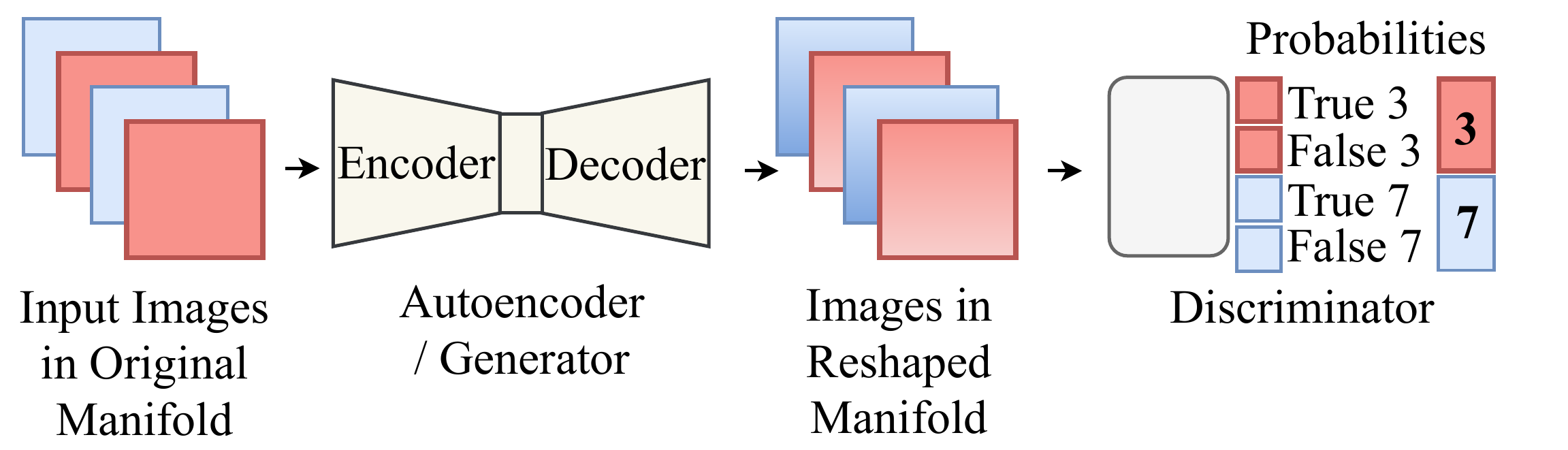}
\end{center}
   \caption{Here, we show an overview of the transformations that images go through in Minimax defense. First, it gets projected on the reshaped manifold by going through the generator. Second, the projected image is classified by the discriminator, where the probability for each label is calculated as the sum of the true and false probabilities for that label.}
\label{fig:chartMinimaxArch}
\end{figure*}

\textbf{Minimax defense architecture.} Our novel Minimax defense against adversarial attacks is a GAN. The generator of this GAN is an autoencoder that takes images as input and outputs images. In addition to the $Real$ and $Fake$ labels of a GAN discriminator, we have added the class labels to the discriminator. As a result, the number of labels in the discriminator is twice the number of labels in the dataset. The architecture of our Minimax GAN is shown in Figure~\ref{fig:chartMinimaxArch}.

Figure~\ref{fig:chartMinimaxArch} also shows how an image in a digit binary classification problem ($3$ and $7$ for simplicity) gets classified by Minimax defense. First, the image goes through the generator and then through the discriminator. The probability of the input image being, for example, digit $3$ is the sum of the probabilities of labels $Real-3$ and $Fake-3$.

\textbf{GAN Discriminators} The discriminators for all three datasets have convolutional layers, batch-normalization layers, max-pooling, drop-out layers. They all use regularization and SGD optimization. The architecture of all three discriminators is shown in Table~\ref{table-discriminators}.

\begin{table}
\begin{center}
\begin{tabular}{|l|l|l|} 
 \hline
 MNIST & CIFAR & TRAFFIC \\
\hline
 Conv.R 3x3x32   & Conv.E 3x3x32  & Conv.R 3x3x32 \\ 
 BatchNorm     & BatchNorm    & BatchNorm   \\
 Conv.R 3x3x64   & Conv.E 3x3x32  & Conv.R 3x3x32 \\
 BatchNorm     & BatchNorm    & BatchNorm   \\
 MaxPool 2x2   & MaxPool 2x2  & MaxPool 2x2 \\
 Dropout 0.25  & Dropout 0.2  & Dropout 0.2 \\
 Dense 128     & Conv.E 3x3x64  & Conv.R 3x3x64 \\ 
 Dropout 0.5   & BatchNorm    & BatchNorm   \\
 Dense.Soft 20 & Conv.E 3x3x64  & Conv.R 3x3x64 \\
               & BatchNorm    & BatchNorm   \\
               & MaxPool 2x2  & MaxPool 2x2 \\
               & Dropout 0.3  & Dropout 0.2 \\
               & Conv.E 3x3x128 & Conv.R 3x3x128 \\ 
               & BatchNorm    & BatchNorm    \\
               & Conv.E 3x3x64  & Conv.R 3x3x128 \\
               & BatchNorm    & BatchNorm    \\
               & MaxPool 2x2  & MaxPool 2x2  \\
               & Dropout 0.4  & Dropout 0.2   \\ 
               & Dense.Soft 20     & Dense 512     \\
               &              & BatchNorm     \\
               &              & Dropout 0.5    \\
               &              & Dense.Soft 20       \\
\hline 
\end{tabular}
\end{center}
\caption{Discriminator architecture for MNIST, CIFAR and TRAFFIC datasets.}
\label{table-discriminators}
\end{table}

\textbf{GAN Generators.} The generators for all three datasets have convolutional layers, batch-normalization layers, max-pooling, drop-out layers. The final layer is a convolutional layer with sigmoid activation. They all use regularization and Adadelta optimization. The architecture of all three generators is shown in Table~\ref{table-generators}.

\begin{table}
\begin{center}
\begin{tabular}{|l|l|l|} 
 \hline
 MNIST          & CIFAR & TRAFFIC \\
\hline
 Conv.R 3x3x32    & Conv.R 3x3x64    & Conv.R 3x3x32 \\ 
 BatchNorm      & BatchNorm      & BatchNorm   \\
 MaxPool 2x2    & Conv.R 3x3x64    & MaxPool 2x2 \\
 Conv.R 3x3x32    & BatchNorm      & Conv 3x3x32 \\
 BatchNorm      & MaxPool 2x2    & BatchNorm   \\
 MaxPool 2x2    & Conv.R 3x3x128   & MaxPool 2x2 \\
 Conv.R 3x3x32    & BatchNorm      & Dropout 0.2 \\ 
 BatchNorm      & Conv.R 3x3x128   & Conv.R 3x3x32   \\
 Upsampling 2x2 & BatchNorm      & BatchNorm \\
 Conv.R 3x3x32    & MaxPool 2x2    & Upsampling 2x2 \\ 
 BatchNorm      & Upsampling 2x2 & Conv 3x3x32   \\
 Upsampling 2x2 & Conv.R 3x3x128   & BatchNorm \\ 
 Conv.Sig 3x3x1     & BatchNorm      & Upsampling 2x2 \\
                & Upsampling 2x2 & Dropout 0.2 \\
                & Conv.R 3x3x64    & Conv.Sig 3x3x3 \\
                & BatchNorm & \\ 
                & Conv.Sig 3x3x3 & \\
\hline 
\end{tabular}
\end{center}
\caption{Generator architecture for MNIST, CIFAR and TRAFFIC datasets.}
\label{table-generators}
\end{table}

\section{Experiments and Results}

To evaluate our Minimax defense method, it is important to show that the accuracy of adversarial samples with Minimax is comparable to the accuracy of non-adversarial samples with a default classifier. We present the results first followed by the details for obtaining them.

\textbf{Results of Minimax defense evaluation.} Our results in Table~\ref{table-mnist-no-adv} show that our Minimax adversarial defense maintains very high accuracy for the MNIST dataset, with values very close to the accuracy of the default classifier trained and tested on the original MNIST dataset - 98.93\%.

\begin{table}[h!]
\begin{center}
\begin{tabular}{|l|l|r|r|} 
 \multicolumn{4}{c}{MNIST Default Accuracy \textbf{98.93\%}} \\
 \hline
 Attack    & Parameter & No       & Minimax \\
           &           & Defense  & Defense \\
 \hline

 
 FGSM           & $\epsilon=0.1$ & 81.18\% & 97.01\%  \\
 \hline
 CW\_L2         & $conf=40$ & 0.79\% & 97.50\%  \\
 \hline
 CW\_L2         & $conf=0$ & 0.84\% & 98.07\%  \\
 \hline
 DeepFool       &           & 1.12\% & 98.87\%  \\
 \hline
\end{tabular}
\end{center}
\caption{Comparison of MNIST dataset classification accuracy for no defense and for Minimax defense. MNIST default accuracy is at the top of the table.}
\label{table-mnist-no-adv}
\end{table}

Evaluating our Minimax adversarial defense on CIFAR-10, our results in Table~\ref{table-cifar-no-adv} show that the Minimax accuracy for the CIFAR-10 dataset remains very close to the accuracy of the default classifier trained and tested on the original CIFAR-10 dataset - 83.14\%. 

\begin{table}[h!]
\begin{center}
\begin{tabular}{|l|l|r|c|}
 \multicolumn{4}{c}{CIFAR Default Accuracy \textbf{83.14\%}} \\
 \hline
 Attack    & Parameter       & No       & Minimax \\
           &                 & Defense  & Defense \\
 \hline
 FGSM     & $\epsilon=0.1$ & 10.28\% & 76.79\% \\
 CW\_L2     &   $conf=40$    &  8.82\% & 69.73\% \\
 CW\_L2     &   $conf=0$     &  8.73\% & \textit{73.90\%} \\
 DeepFool &                &  8.99\% & \textit{76.61\%} \\
 \hline
\end{tabular}
\end{center}
\caption{We compare CIFAR-10 dataset classification accuracy for no defense and for Minimax defense. \textit{Adversarial training} was used for the results in italics.}
\label{table-cifar-no-adv}
\end{table}

We evaluate our Minimax adversarial defense on TRAFFIC and find, based on in Table~\ref{table-traffic-no-adv}, that the accuracy of our Minimax adversarial defense for the TRAFFIC dataset remains very close to the accuracy of the default classifier trained and tested on the original TRAFFIC dataset - 96.97\%.

\begin{table}[h!]
\begin{center}
\begin{tabular}{|l|l|r|r|}
 \multicolumn{4}{c}{TRAFFIC-32x32 Default Accuracy \textbf{96.97\%}} \\
 \hline
 Attack    & Parameter       & No      & Minimax \\
           &                 & Defense & Defense \\
 \hline
 FGSM     & $\epsilon=0.1$  & 28.74\%  & 81.41\% \\
 CW\_L2     &   $conf=40$     &  1.56\%  & \textit{94.54\%} \\
 CW\_L2     &   $conf=0$      &  1.41\%  & \textit{93.66\%} \\
 DeepFool &                 & 1.43\%   & 94.57\% \\
 \hline
\end{tabular}
\end{center}
\caption{We compare classification accuracy for the TRAFFIC dataset for no defense and for Minimax defense. \textit{Adversarial training} was used for the results in italics.}
\label{table-traffic-no-adv}
\end{table}

\textbf{Comparison with MagNet.} The MagNet method has been shown~\cite{carlini2017magnet} to not withstand customized CW attacks. Nevertheless, we compare the two approaches to provide context for our results since they both use autoencoders.

\begin{table}[h!]
\begin{center}
\begin{tabular}{|l|l|r|r|}
 \multicolumn{3}{c}{Comparison to MagNet for MNIST} \\
 \hline
 Attack    & Minimax / No     & MagNet   / No     \\
           & defense / attack & Defense  / attack \\
 \hline
 FGSM $\epsilon=0.3$   & \textbf{86.72\%} / 98.93\% & \textbf{*0.13\%} / 99.4\% \\
 \hline
 FGSM $\epsilon=0.2$   & 98.52\% / 98.93\% & NA \\
 \hline
 FGSM $\epsilon=0.1$   & 98.52\% / 98.93\% & NA \\
 \hline
 FGSM $\epsilon=0.01$  & 98.52\% / 98.93\% & 100.00\% / 99.4\% \\
 \hline
 CW\_L2     & 97.50\% / 98.93\% & 99.50\% / 99.4\% \\
 $conf=40$  &          & \\
 \hline
 CW\_L2     & 98.07\% / 98.93\% & 99.50\% / 99.4\% \\
 $conf=0$   &          & \\
 \hline
 DeepFool   & 98.87\% / 98.93\% & 99.40\% / 99.4\% \\
 \hline
 
\end{tabular}
\end{center}
\caption{Comparison of Minimax defense results to MagNet results. The Magnet results are obtained from the Magnet paper~\cite{meng2017magnet} with the exception of entries marked with *, which were averaged from comparison results from the Defense-GAN~\cite{samangouei2018defense} paper.}
\label{comparison-magnet}
\end{table}

MagNet~\cite{meng2017magnet} is tested against very small perturbations for FGSM attack on MNIST dataset - up to $\epsilon=0.01$, whereas we show Minimax withstands attacks of very high perturbations, up to $\epsilon=0.3$, shown here in Table~\ref{comparison-magnet}.

\textbf{Attacks.} We generate adversarial samples with CleverHans 3.0.1~\cite{papernot2018cleverhans} for the FGSM, CW\_L2 and DeepFool attacks and the IBM Adversarial Robustness 360 Toolbox (ART) toolbox~\cite{art2018} for the JSMA attack. We evaluate against three types of white-box attacks: CW~\cite{carlini2017towards}, DeepFool~\cite{moosavi2016deepfool} and FGSM~\cite{goodfellow6572explaining}. We choose CW~\cite{carlini2017towards} and DeepFool~\cite{moosavi2016deepfool} because they are state-of-the-art, and FGSM~\cite{goodfellow6572explaining} because it is one of the first identified attacks. Minimax is written in Python 3.5.2, using Keras 2.2.4~\cite{chollet2015keras}.

\textbf{Baselines.} To calculate a baseline in the absence of adversarial attack, we trained classifiers with identical models as our Minimax discriminators, except for the number of output labels in the last softmax layer. The baselines in the absence of adversarial attack are: for MNIST 98.93\%, for CIFAR-10 83.14\%, for TRAFFIC 96.97\%.

\textbf{Datasets} We evaluate our Minimax adversarial defense on three datasets - MNIST~\cite{lecun1998mnist}, CIFAR-10~\cite{krizhevsky2009cifar} and TRAFFIC~\cite{Stallkamp2012}. The MNIST dataset~\cite{lecun1998mnist} is a dataset of hand-written digits with ten classes of size $28 \times 28 \times 1$. MNIST has 60K training samples and 10K testing samples. The CIFAR-10 dataset~\cite{krizhevsky2009cifar} is a 10-class dataset of objects, with images of size $32 \times 32 \times 3$. The sizes of the CIFAR-10 training and testing datasets are 50K and 10K. The TRAFFIC dataset~\cite{Stallkamp2012} is a dataset of 43 classes of images of traffic signs in different sizes, from $15 \times 15 \times 3$ to $250 \times 250 \times 3$ pixels. We rescale the images to $32 \times 32 \times 3$ the same size as CIFAR-10 images. The TRAFFIC training and testing datasets have respectively 39209 and 12630 samples.

\textbf{Discriminator and generator architectures.} The model architecture of all discriminators is shown in Table~\ref{table-discriminators}, the model architecture of all autoencoders is shown in Table~\ref{table-generators}.

\textbf{Optimizers.} For classification of the MNIST dataset, we use an Adadelta optimizer for the discriminator, classifier and generator; and SGD for the adversarial model. For CIFAR-10, we use RMSprop for the discriminator and classifier; Adadelta for the generator; and SGD for the adversarial model. For the TRAFFIC dataset, we use SGD for the discriminator and classifer; Adadelta for the generator; SGD for the adversarial model.

\textbf{Training.} We use $L2$ kernel regularization in the convolutional layers of the discriminators, classifiers, and generators for the classification of all three datasets. We do not perform image augmentation in the training of any of the datasets. The batches are chosen randomly. The training is done in random batches of 32 images for MNIST and TRAFFIC, and 128 images for CIFAR-10. The number of epochs is 10 for MNIST, TRAFFIC, and 64 for CIFAR-10. We conduct experiments with and without adversarial training. For adversarial training, we enhance the training dataset with adversarial samples of the same type as the the attack. During training, the training adversarial samples are refreshed after 60 batches.

 
\section{Discussion and Conclusions}

In this paper, we have shown that all state-of-the-art attacks were gradient-based and that our Minimax defense countered state-of-the-art adversarial attacks against neural-network classifiers. This was done without masking gradients which has been shown to be an ineffective defense. Since all state-of-the-art gradient attacks are gradient-based, minimax defends against all of them. Our novel Minimax defense has combined a minimax approach against gradient-based attacks with an approach that reshapes the manifold for better classification. Our results showed that Minimax defense countered gradient-based attacks for three diverse datasets. For CW attacks, Minimax defense achieved 98.07\% (MNIST-default 98.93\%), 73.90\% (CIFAR-10-default 83.14\%) and 94.54\% (TRAFFIC-default 96.97\%). Against DeepFool attacks, our minimax defense achieves 98.87\% (MNIST), 76.61\% (CIFAR-10) and 94.57\% (TRAFFIC). These results show that Minimax maintains accuracy for adversarial samples.

We have also demonstrated using the TRAFFIC dataset in adversarial attacks and our Minimax defense. The TRAFFIC dataset could be a replacement for CIFAR-10 in adversarial methods. Though CIFAR-10 is commonly used for adversarial attacks and defenses, it needs many epochs to converge.

In conclusion, we have identified gradient descent as an underlying crucial aspect of current attacks, which has lead to our Minimax defense that does not mask its gradients but is still able to counter state-of-the-art attacks.






{\small
\bibliographystyle{ieee_fullname}
\bibliography{egbib}
}

\end{document}